\pdfoutput=1

\documentclass[11pt]{article}

\usepackage[final]{acl}

\usepackage{times}
\usepackage{latexsym}

\usepackage{soul}
\usepackage{times}
\usepackage{empheq}
\usepackage{latexsym}
\usepackage{booktabs}
\usepackage{multirow}
\usepackage{subfigure}
\usepackage{tabularx}
\usepackage{color}
\usepackage{graphicx}
\usepackage{caption}
\usepackage{amsmath}
\usepackage{placeins}
\usepackage{colortbl}
\usepackage{needspace}

\usepackage[ruled,vlined]{algorithm2e}
\usepackage{algorithmic}
\usepackage{float}
\usepackage{graphicx}
\usepackage{subcaption}

\usepackage[T1]{fontenc}

\usepackage[utf8]{inputenc}

\usepackage{microtype}

\usepackage{inconsolata}

\usepackage{graphicx}

\newcommand{\inst}[1]{\textsuperscript{#1}}

\title{Explicit and Implicit Data Augmentation for Social Event Detection}

\author{Congbo Ma\inst{1,2,3}, Yuxia Wang\inst{2}, Jia Wu\inst{1}, Jian Yang\inst{1}, Jing Du\inst{1}, \\
\textbf{Zitai Qiu\inst{1}}, \textbf{Qing Li\inst{2}}, \textbf{Hu Wang\inst{2}}, \textbf{Preslav Nakov\inst{2}} \\
\inst{1} Macquarie University, Sydney, Australia\\
\inst{2}Mohamed bin Zayed University of Artificial Intelligence, UAE \\
\inst{3}New York University Abu Dhabi, UAE \\
}

\begin{document}
\maketitle
\begin{abstract}
Social event detection involves identifying and categorizing important events from social media, which relies on labeled data, but annotation is costly and labor-intensive. To address this problem, we propose \textbf{Aug}mentation framework for \textbf{S}ocial \textbf{E}vent \textbf{D}etection (\textbf{SED-Aug}), a plug-and-play dual augmentation framework, which combines explicit text-based and implicit feature-space augmentation to enhance data diversity and model robustness.
The explicit augmentation utilizes large language models  to enhance textual information through five diverse generation strategies. For implicit augmentation, we design five novel perturbation techniques that operate in the feature space on structural fused embeddings. These perturbations are crafted to keep the semantic and relational properties of the embeddings and make them more diverse. Specifically, SED-Aug outperforms the best baseline model by approximately 17.67\% on the Twitter2012 dataset and by about 15.57\% on the Twitter2018 dataset in terms of the average F1 score. The code is available at GitHub\footnote{https://github.com/congboma/SED-Aug}.
\end{abstract}

\section{Introduction}

Social event detection (SED) identifies and classifies notable events on social media platforms \cite{hao2021streaming, Cao2023Hierarchical, Ren2022Evidential, Cao2021Knowledge, ma2025enhanced}. These events, distinct from general occurrences, are characterized by their origin and propagation through user interactions, reflecting collective activity. 
SED is commonly defined as a classification task that analyzes social network data, including textual content (e.g., messages) and structural features (e.g., user metadata and activity logs), to detect emerging events. These insights can be important for various applications such as crisis management \cite{pekar2020early}, public opinion analysis\cite{peng2021streaming}, and financial market analysis \cite{nisar2018twitter}. However, one key challenge in SED is its reliance on limited labeled data, which demands human annotation and hampers the generalization of models to diverse event contexts. \cite{qiu2024heterogeneous}.

Large language models (LLMs) excel in natural language understanding and generation, enabling the creation of diverse textual variations. Despite their potential, LLMs have not yet been applied to the SED task. When used for textual augmentation, LLMs generate  social message variations, enriching SED training data and improving model robustness. This approach also reduces computational overhead during testing by handling intensive computations in advance.
Once data are augmented, the event detection can be conducted without invoking the LLMs, thus eliminating recurring time costs associated with model inference (usually, LLMs inference will be time consuming). Moreover, Direct LLM-based predictions are expensive due to API or cloud fees. Using LLMs solely for augmentation enhances data diversity while optimizing computational and financial resources.

Focusing solely on textual data overlooks crucial structural information that captures user and event interactions in social media \cite{qiu2024heterogeneous}. Common SED methods use graph-based approaches to generate structure-fused embeddings \cite{Cao2021Knowledge, peng2022reinforced, Ren2022From, qiu2024heterogeneous, li2024relational, ma2024learning}.
However, while LLMs excel at augmenting textual data, they are less effective at processing graph-based data \cite{jin2024large}. To address this, we extend data augmentation to structural domain, which captures both the information of social messages and the relational patterns among associated metadata. This helps incorporate the structural information and diversify the data from a different perspective to further improve SED performance.

\needspace{1\baselineskip}
In this paper, we propose \textbf{Aug}mentation framework for \textbf{S}ocial \textbf{E}vent \textbf{D}etection (\textbf{SED-Aug}), a dual plug-and-play data augmentation framework for SED. It includes explicit augmentation using LLMs to enhance textual diversity, and implicit augmentation, which perturbs structure-fused embeddings in the feature space.
Explicit augmentation consists of one-stage and two-stage strategies. One-stage includes paraphrasing, context addition, style transfer, and paraphrasing with entity preservation. Two-stage first extracts key information using LLMs, then rewrites it into diverse messages. These strategies enable the generation of a wide range of augmented messages, increasing the variability and robustness of the training data.
Implicit augmentation introduces five novel perturbation techniques that operate directly in the feature space of structure-fused message embeddings. These methods include Gaussian Perturbation (GP), Proportional Gaussian Perturbation (PGP), In-Distribution Gaussian Perturbation (IDGP), Clipped Gaussian Perturbation (CGP), and Frequency-Domain Perturbation (FDP). Each perturbation is designed to modify the embeddings while preserving their semantic and relational properties, ultimately enhancing the model to capture intricate patterns within data.

This dual augmentation framework leverages both the textual and structural information inherent in SED, effectively enhancing data diversity and capturing more complex patterns within the underlying social graph. By addressing the unique characteristics of SED, it maximizes the potential of both semantic and structural embeddings, leading to more robust and accurate detection outcomes.
The primary contributions of our work can be summarized in the following key aspects: 

\begin{itemize}
\vspace{-1mm}
\item We propose a novel dual data augmentation framework SED-Aug tailored for SED, integrating both explicit and implicit augmentation strategies. This framework is plug-and-play and can be seamlessly integrated into a SED model to enhance its performance and robustness.
\vspace{-1mm}
\item We are the first to incorporate LLMs in addressing the SED task, developing multiple augmentation strategies for message content. These include five one-stage and two-stage techniques that increase data diversity without the need for manual labeling.
\vspace{-1mm}
\item We propose five perturbation methods for the feature space, which operate on structural fused message embeddings. This integration of structural information into the augmentation process enhances the robustness and performance of SED models.
\vspace{-1mm}
\item The proposed augmentation framework achieves state-of-the-art results on several SED datasets, outperforming the best baseline model by approximately 17.67\% on the Twitter2012 dataset and about 15.57\% on the Twitter2018 dataset in terms of average F1 score, supported by comprehensive analysis validating its effectiveness.

\end{itemize}

\begin{figure*}[t]
\centering
\includegraphics[width=0.9\textwidth]{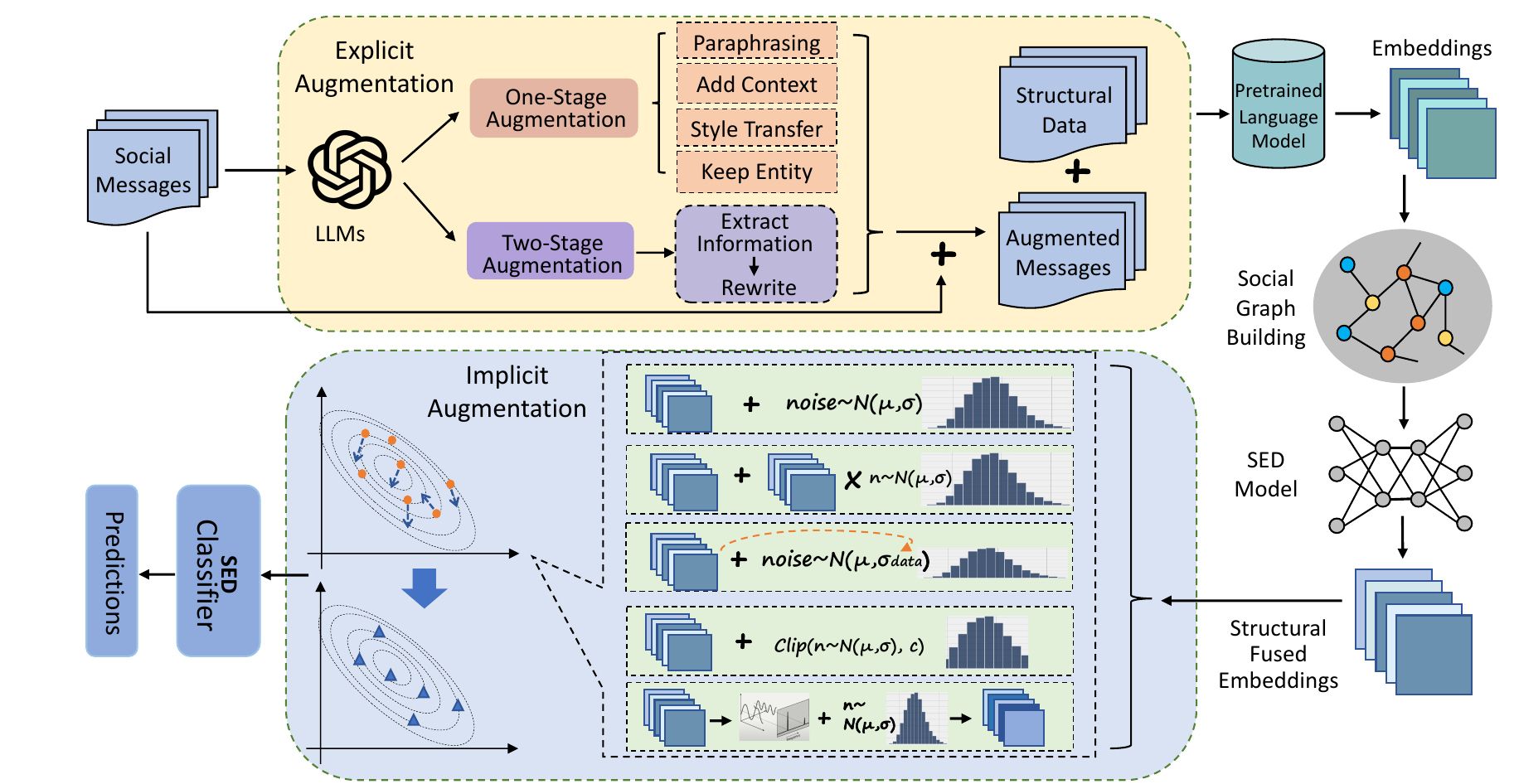}
\vspace{-1mm}
\caption{The framework of SED-Aug model.}
\label{fig:framework}
\end{figure*}

\section{Related Works}
\subsection{Social Event Detection}
Early SED research primarily used content-based methods focusing solely on text semantics \cite{Wurzer2015Twitter, Wang2017A, Yan2015A}, neglecting the importance of social interactions and the heterogeneous nature of social media data, such as user connections, interactions, and spatiotemporal information \cite{toivonen2019social, marti2019social}. As a result, this limitation restricts the ability to capture a comprehensive range of information, leading to missed insights from diverse data types \cite{Ren2022Evidential}.

Graph-based methods \cite{Kipf2017Semi, Velickovic2018Graph, Hamilton2017Inductives, wu2020comprehensive, zhang2019heterogeneous} address this by leveraging both textual and structural data through heterogeneous information networks \cite{sun2012mining}, modeling complex social media interactions \cite{Cao2021Knowledge, peng2022reinforced, Ren2022From, qiu2024heterogeneous, li2024relational}.
However, these methods do not fully leverage existing data to enhance data diversity, overlooking the potential of data augmentation techniques to improve model performance, particularly in situations with limited labeled data.

\subsection{Data Augmentation}

Data augmentation is crucial in natural language processing (NLP) for enhancing textual diversity and improving model generalization, particularly when labeled data is limited \cite{feng2021survey}. It introduces variability to training data and can be broadly categorized into two types: (1) input text level augmentation and (2) feature space level augmentation.

For (1), various token and sentence level modifications diversify training data. Common strategies include insertion \cite{Xie2020Unsupervised}, deletion \cite{Jason2019EDA}, and masking \cite{Sreyan2023A, Yu2023Cross} to improve generalization. Replacement techniques \cite{Sosuke2018Contextual} include synonym substitution, entity replacement \cite{liu2023improving}, and semantic modifications \cite{zhuang2022learning}. Other approaches include paraphrasing and back-translation \cite{Sennrich2016Improving} for generating semantically equivalent variations.

For (2), augmentation transforms feature representations \cite{Terrance2017Dataset}. For example, Ang et al. \cite{Ang2023DialoGPS} modeled dialogue trajectories using a Gaussian process, while Wang et al. \cite{Wang2019Implicit} estimated class-wise covariance matrices to generate synthetic data, optimizing cross-entropy loss. However, these methods focus on textual and overlook structural information, limiting their effectiveness in social network analysis.
Unlike prior work that treats textual and structural augmentation separately, we propose SED-Aug, a dual data augmentation framework for SED. By integrating explicit and implicit strategies, we enhance data diversity and model robustness in low-resource settings.

\section{Methodology}
In this section, we present the SED-Aug framework, as illustrated in Figure~\ref{fig:framework}. We first apply explicit augmentation to social media messages using LLMs, leveraging five strategies to enhance the user-generated content. We then combine these augmented messages with the original messages and integrate them with the corresponding structural data of each message.
Next, we use a pre-trained language model to extract embeddings from this enriched data. We construct a social graph that captures the connections between the textual content and the structural information. Through a graph aggregation method, we generate structural-fused message embeddings that encapsulate both semantic and relational information.
These structural-fused embeddings are then subjected to implicit augmentation to create more diverse embeddings in the feature space. Finally, we integrate these augmented embeddings with the original structural-fused embeddings for the downstream classification task, improving the robustness and adaptability of the SED model.

\subsection{Explicit Data Augmentation through LLMs}

We introduce five explicit augmentation techniques using LLMs to enhance SED. These methods modify the message text while preserving metadata to create diverse messages that can enhance the robustness and generalization capabilities of models for SED. Given an input message:  

\vspace{-2mm}
\begin{equation} \small
    m_i^{k} = LLM^k(m_{i})
\end{equation}

\noindent where \( k \) represents different augmentation types. The five methods are described as follows:

\noindent \textbf{Paraphrasing.} Generating alternate versions of the original message while preserving its meaning, maintaining semantic integrity with variation in wording and structure.

\noindent
\textbf{Adding Context.} Expanding the original message by adding relevant contextual information. This augmentation helps provide a broader understanding of the content, enriching the message with details that enhance its clarity and relevance.

\noindent
\textbf{Style Transfer.} Modifying the writing style of the message without altering its core meaning. The style perturbation can include shifts in tone, formality, or other stylistic attributes to produce varied versions of the same message.

\noindent
\textbf{Keep Entity Unchanged.} Entities are crucial for SED, representing key components like people, places, dates, or topics that provide context and relevance to the message content. Maintaining these entities preserves critical event details, helping the model distinguish and understand events. This approach modifies the message text while ensuring key entities remain unchanged, generating diverse variations while retaining essential information.

\noindent
\textbf{Extract Informative Information and Rewrite.} This method uses an LLM to extract key components from the original message and generate a revised version that preserves essential information while modifying its presentation. We focus on (1) the selection of keywords is determined by the background knowledge of the LLM; (2) given the importance of entities in SED, we specifically guide the LLM to concentrate on entity extraction for more targeted augmentation. The LLM extracts key entities (e.g., names, locations, dates) and integrates them into a restructured message, enhancing diversity while preserving core information; (3) a knowledge graph is a structured framework that captures relationships among entities, providing a clearer understanding of the connections within the data. The LLM rewrites the message using this structured information, enriching the text while preserving key relationships.

\subsection{Implicit Data Augmentation on the Latent Space}

In the latent space, we aim to introduce minor variations that enhance data diversity by applying perturbations to augment the structural fused message embeddings $g^{i}$, where $g^{i} \in G$, $i = 1, 2, \ldots, N$. $G$ is the set of structural fused message embeddings. $N$ is the number of message samples. This augmentation process improves the model's generalization capability. During training, a probability threshold $\alpha$ is employed to determine whether to train on the augmented embeddings or the original embeddings, which can be expressed as follows:
\vspace{-2mm}
\begin{equation} \small
g^i = 
\begin{cases} 
g^i_{\text{augmented}}, & \text{if } p < \alpha, \\
g^i, & \text{if } p \geq \alpha,
\end{cases}
\end{equation}

\vspace{-2mm}
\begin{equation} \small
p \sim \text{Uniform}(0, 1).
\end{equation}

\noindent Here, $p$ is a real number sampled from a uniform distribution. If $p < \alpha$, we train on the augmented embeddings to introduce diversity; otherwise, we use the original embeddings  to maintain stability. We propose five implicit augmentation methods to enrich the latent space.

\noindent\textbf{Gaussian Perturbation (GP).}
For each input structural fused message embeddings $g^{i}$, we add Gaussian noise sampled from a normal distribution with a mean of zero and a standard deviation of $\sigma$. The noise is applied element-wise to each dimension, generating a perturbed version of the original feature vector $g^{i}_{GP}$. Formally, the augmented feature is computed as:

\vspace{-2mm}
\begin{equation} \small
g^{i}_{GP} = g^{i} + n_{GP}, n_{GP} \sim \mathcal{N}(0,\sigma^{2} ),
\end{equation}

\noindent\textbf{Proportional Gaussian Perturbation (PGP). }
In the proposed GP, the scale of the perturbation is independent of the original data scale. This can lead to noise that is disproportionately large or small relative to the magnitude of the input data, potentially distorting the underlying structural information.
To address this, we propose PGP, where the added noise is scaled in relation to the input data in the feature space. Specifically, the noise is generated such that its magnitude is proportional to the values of the feature vector itself. Formally, we have:
\vspace{-2mm}
\begin{equation} \small
g^{i}_{PGP} = g^{i} + n_{PGP}, n_{PGP} \sim \mathcal{N} (0,\sigma^{2}) \cdot G,
\end{equation}

In this perturbation, the noise is not only dependent on the standard deviation but also scales with the absolute value of the original feature vector. This helps that the perturbation is contextually relevant, enhancing the robustness of the feature augmentation process while preserving the intrinsic relationships within the data.

\noindent\textbf{In-Distribution Gaussian Perturbation (IDGP).}
In the GP and PGP, the standard deviation $\sigma$ should be manually predefined, rather than being dynamically adapted based on the distribution of input data. Therefore, we proposed IDGP that the noise is generated based on the statistical properties of the input data, specifically its standard deviation. For each feature dimension, the standard deviation $\sigma$ 
is computed from the structural fused message embeddings $g^{i}$, and Gaussian noise with zero mean and a standard deviation proportional to this $g^{i}$ is added. Formally:
\vspace{-2mm}
\begin{equation} \small
g^{i}_{IDGP} = g^{i} + n_{IDGP}, n_{IDGP} \sim \mathcal{N} (0, \alpha * std(G)^{2} ),
\end{equation}

\noindent where $\alpha$ is the variance control parameters. By adapting the noise magnitude to the inherent variability of the data (through $std(G)$), this perturbation ensures that the noise is scaled appropriately for each feature dimension. This prevents excessive perturbation in low-variance features and ensures sufficient noise is applied in high-variance features. Furthermore, as the perturbations stay within the range of the original data's distribution, this method maintains the data's overall structure and prevents the creation of unrealistic or outlier data points.

\noindent\textbf{Clipped Gaussian Perturbation (CGP). }
This method constrains Gaussian noise within a specified range to maintain balanced perturbation. Noise is sampled from a standard normal distribution and clipped within \(\left[ -c, c \right]\), where \(c\) is a small constant. This prevents the noise from being negligible or excessive, maintaining a balanced perturbation effect. The augmented feature is calculated as follows:

\vspace{-2mm}
\begin{equation} \small
g^{i}_{CGP} = g^{i} + n_{CGP}, \quad \text{Clip}(n_{CGP} \sim\mathcal{N}(0, \sigma^{2}), c), \end{equation}

\noindent CGP transforms its distribution into a truncated normal distribution, characterized by finite bounds that exclude values outside the specified range. While the original noise has tails that extend infinitely, clipping focuses on the central portion of the distribution, resulting in a noise distribution with higher density near the mean and bounded at the edges.

\noindent\textbf{Frequency-Domain Perturbation (FDP).}
To enhance message embedding features, we propose FDP, which applies a Fourier transform to the embeddings, adds noise in the frequency domain, and then transforms the data back to the time domain. This process introduces controlled perturbations to capture diverse embedding characteristics, improving model robustness and generalization.
The process begins by converting \( g^i \) from the time domain to the frequency domain using the Fourier transform:
\vspace{-2mm}
\begin{equation} \small
F^i = \mathcal{F}(g^i),
\end{equation}

\noindent where \( \mathcal{F} \) denotes the Fourier transform, and \( F^i \) represents the transformed embedding in the frequency domain.
We selectively use the frequency components based on a specified keep ratio \( r \). For the high mode, we retain the high-frequency components while attenuating the low-frequency components:
\vspace{-2mm}
\begin{equation} \small
F^i_{filtered} = F^i[N-r*N:N],
\end{equation}

\noindent where $N$ is the number of message sample.
We then add noise directly in the frequency domain to the filtered embedding:
\vspace{-2mm}
\begin{equation} \small
F^i_{FDP} = F^i_{filtered} + n,
n \sim \left( \mathcal{N}(0, \sigma^2) + i \cdot \mathcal{N}(0, \sigma^2) \right) \cdot \eta,
\end{equation}

\noindent $\eta$ is a noise level. We apply the inverse Fourier transform to convert the perturbed frequency-domain representation back into the time domain:
\vspace{-2mm}
\begin{equation} \small
g^i_{FDP} = \mathcal{F}^{-1}(F^i_{FDP}),
\end{equation}

By applying the Fourier transform, this method enables controlled perturbations in the frequency domain, allowing for a targeted enhancement of specific frequency components, which leads to a more nuanced and robust representation of the embedding's features.

\section{Experiments}

The research questions are:
\textbf{Q1:} How does the proposed dual augmentation framework compare to strong baseline models? 
\textbf{Q2:} What are the individual effects of explicit and implicit data augmentation on model performance?
\textbf{Q3:} Which types of extracted information are most effective for rewriting in explicit augmentation?  
\textbf{Q4:} How do different frequency-domain perturbations (e.g., high-frequency noise, band filter, low-frequency noise) impact the model?  
\textbf{Q5:} How do the augmentation methods perform with limited data?
\textbf{Q6:} How does implicit augmentation alter the data distribution?

\begin{table*}[h]
\setlength\tabcolsep{3pt}
    \centering
    \caption{Overall performance comparison, with all results presented as percentages. The improvement represents the percentage increase of our method over the best baseline model.}
    \vspace{-2mm}
    \scalebox{0.7}{\begin{tabular}{@{}l|ccc|ccc|ccc@{}}
\hline
\hline
Datasets &
  \multicolumn{3}{c|}{Kawarith6} &
  \multicolumn{3}{c|}{Twitter2012} &
  \multicolumn{3}{c}{Twitter2018} \\ \hline
Models &
  Micro F1 &
  \cellcolor [HTML]{FFFFFF} Macro F1 &
  Average &
  Micro F1 &
  Macro F1 &
  \cellcolor [HTML]{FFFFFF} Average &
  Micro F1 &
  Macro F1 &
  \cellcolor [HTML]{FFFFFF} Average \\ \hline
TF-IDF   & 92.87 & 92.49 & 92.68 & 67.89 & 34.05 & 50.97 & 42.59 & 20.00 & 31.30 \\
Word2Vec & 65.64 & 56.23 & 60.94 & 57.14 & 28.13 & 42.64 & 53.77 & 22.24 & 38.01 \\
FastText & 85.12 & 82.06 & 83.59 & 17.25 & 0.70  & 8.98  & 1.06  & 0.56  & 0.81  \\
FinEvent & 92.59 & 91.36 & 91.96 & -     & -     & -     & -     & -     & -     \\
BERT     & 76.95 & 75.02 & 75.99 & 68.89 & 51.58 & 60.24 & 55.45 & 32.00 & 43.73 \\
GraphMSE & 94.70 & 94.00 & 94.35 & 80.57 & 67.16 & 73.87 & 76.71 & 66.21 & 71.46 \\
ETGNN    & -     & -     & -     & 84.80 & 75.65 & 80.23 & -     & -     & -     \\
HGT      & 91.93 & 91.18 & 91.56 & 71.11 & 58.41 & 64.76 & 80.46 & 68.92 & 74.69 \\
KPGNN    & 78.63 & 76.91 & 77.77 & -     & -     & -     & -     & -     & -     \\
GraphHAM & 95.10 & 94.57 & 94.84 & 84.14 & 71.00 & 77.57 & 80.54 & 71.77 & 76.16 \\
 \hline
SED-Aug  & \textbf{98.41} & \textbf{98.29} & \textbf{98.35} & \textbf{93.03} & \textbf{89.53} & \textbf{91.28} & \textbf{89.61} & \textbf{86.43} & \textbf{88.02} \\ 
Improvement  &3.48 $\uparrow$ 	&3.93 $\uparrow$	&3.70 $\uparrow$	&10.57 $\uparrow$	&26.10 $\uparrow$	&17.67 $\uparrow$	&11.26 $\uparrow$	&20.43 $\uparrow$	&15.57 $\uparrow$\\\hline
\hline
\end{tabular}}
\label{tab:overall}
\end{table*}

\subsection {Dataset, Evaluation Metrics and Baselines}
We conduct experiments on three datasets:  Kawarith6 \cite{alharbi2021kawarith}, Twitter2012 \cite{James2013Building}, and Twitter2018\cite{Beatrice2020a}. 
For evaluation, we use Micro F1 and Macro F1. Micro F1 evaluates the overall performance in classifying instances across all classes, while Macro F1 checks each class individually before averaging these scores.
The baseline models we have compared with are: TF-IDF \cite{akiko2003an}, Word2Vec \cite{mikolov2013efficient}, FastText \cite{armand2017bag}, FinEvent \cite{peng2022reinforced}\footnote{We  also tried FinEvent on Twitter2012 and Twitter2018, but it adopts GAT for message passing that is very memory-intensive and casted out-of-memory errors.}, BERT \cite{kenton2019bert}, GraphMSE \cite{li2021graphmse}, ETGNN \cite{Ren2022Evidential}\footnote{We report the performance of ETGNN from the paper~\cite{Ren2022Evidential}, they evaluated the model on Kawarith7, which have different data scales with ours.}, KPGNN \cite{Cao2021Knowledge}\footnote{KPGNN encounter OOM errors on the Twitter2012 and Twitter2018, and thus their results are not included.}, HGT \cite{Hu2020Heterogeneous}, and GraphHAM \cite{Qiu2024an}.

\vspace{-1mm}

\subsection{Overall Performance (Q1)}
The experimental results presented in Table \ref{tab:overall}, highlighting the overall performance comparison of various models. the SED-Aug model demonstrated state-of-the-art performance across all datasets, showing clear improvements in both Micro F1 and Macro F1 metrics, indicating its robustness and effectiveness in handling diverse SED tasks.
Specifically, on the Kawarith6 dataset, SED-Aug achieved a Micro F1 score of 98.41\%, a Macro F1 score of 98.29\%, and an average of 98.35\%, surpassing the performance of the next best models, GraphHAM, which had averages of 94.84\%.
Our proposed model, SED-Aug, achieved notable improvements on both the Twitter2012 and Twitter2018 datasets. On the Twitter2012 dataset, SED-Aug showed a 10.57\% increase in Micro F1 and a 26.10\% increase in Macro F1 compared to the best baseline model. Similarly, on the Twitter2018 dataset, SED-Aug demonstrated an improvement of 11.26\% in Micro F1 and 20.43\% in Macro F1 over the best-performing baseline. These results clearly demonstrate the effectiveness of our dual data augmentation approach in enhancing the performance for SED.

\begin{table}[h]
\setlength\tabcolsep{3pt}
\centering
\caption{Different explicit augmentation methods, with all results presented as percentages.}
\vspace{-2mm}
\scalebox{0.6}{\begin{tabular}{@{}l|cc|cc|cc@{}}
\hline
\hline
Datasets & \multicolumn{2}{c|}{Kawarith6} & \multicolumn{2}{c|}{Twitter2012} & \multicolumn{2}{c}{Twitter2018} \\ \hline
Methods          & Micro F1 & Macro F1 & Micro F1 & Macro F1 & Micro F1 & Macro F1 \\\hline
Paraphrase       & 97.69    & 97.35    & 92.70    & \textbf{89.68}    & 87.64    & 81.68    \\
Style transfer   & 97.12    & 96.69    & 91.60    & 86.92    & 86.33    & 79.29    \\
Add context      & 97.94    & 97.75    & 91.74    & 87.19    & 87.02    & 79.85    \\
Keep entity      & \textbf{98.20}     & \textbf{98.08}    & \textbf{92.76}    & 89.33    & \textbf{88.70}     & \textbf{82.14}    \\
Extract rewrite & 97.99    & 97.84    & 92.34    & 88.65    & 88.47    & 81.57    \\ \hline \hline
\end{tabular}}
\label{tab:Explicit_aug}
\end{table}

\begin{table}[h]
\centering
\caption{Performance of combining the most and least effective explicit augmentations with all five implicit augmentations on the Twitter2012 dataset.}
\vspace{-2mm}
\scalebox{0.6}{
\begin{tabular}{l|c|c|c|c}
\hline \hline
Twitter2012 & \multicolumn{2}{c|}{Keep Entity} & \multicolumn{2}{c}{Style Transfer} \\
\hline
Methods & Micro F1 & Macro F1 & Micro F1 & Macro F1 \\
\hline
GP   & 91.83 & 87.16 & 92.91 & 89.47 \\
PGP  & 91.73 & 86.97 & 92.85 & 89.65 \\
IDGP & 91.79 & 86.93 & 93.01 & 89.58 \\
CGP  & 91.70 & 87.15 & 93.03 & 89.53 \\
FDP  & 91.81 & 87.19 & 92.89 & 89.96 \\
\hline \hline
\end{tabular}}
\label{tab:exp_imp_twitter2012}
\end{table}

\subsection{Ablation Studies (Q2)}

Table \ref{tab:Explicit_aug} shows that all five explicit augmentation methods improve the performance of the SED model. Among them, the "keep entity" method consistently performs best across all datasets, achieving the highest Micro F1 and Macro F1 scores on both Kawarith6 and Twitter2018, and the highest Micro F1 and second-highest Macro F1 on Twitter2012.  Preserving these entities ensures that the model retains essential context, enabling it to accurately identify and detect significant events.
We acknowledge that using LLMs to add context may introduce hallucinations. However, the brevity of our additions limits the risk of false information. Studies show that GPT-4 generates about 6\% false claims in long-form responses \cite{Wang2024OpenFactCheck}. To ensure quality, we sampled 50 examples and found false information in 3, indicating an acceptable factuality level for this augmentation strategy.

To explore the interactions between explicit and implicit augmentations, we further selected a subset of combinations from the 75 possible cases (5 explicit methods × 5 implicit methods × 3 datasets) based on the individual effectiveness of each augmentation. Specifically, we selected the most and least effective explicit augmentation among the five and combined each with all five implicit augmentations to analyze their interactions. The results \ref{tab:exp_imp_twitter2012} show that implicit augmentation consistently provides additional benefits when combined with explicit augmentation, with no observed cases of performance degradation.

Using "keep entity" as the baseline, we then evaluate different implicit augmentation strategies in all three datasets. Table \ref{tab:Implicit_aug} shows that all five methods improve Micro and Macro F1 scores in most cases, with variations across datasets. PGP achieved the highest F1 scores on both the Kawarith6 and Twitter2018 datasets, attributed to its method of scaling the added perturbation relative to the input data's feature values, ensuring that the perturbations remain contextually relevant to the data's magnitude. CGP attains the best Micro F1 on Twitter2012 by clipping extreme noise values, preventing outliers and preserving feature distribution.

\begin{figure}[t]
\centering
\includegraphics[width=0.48\textwidth]{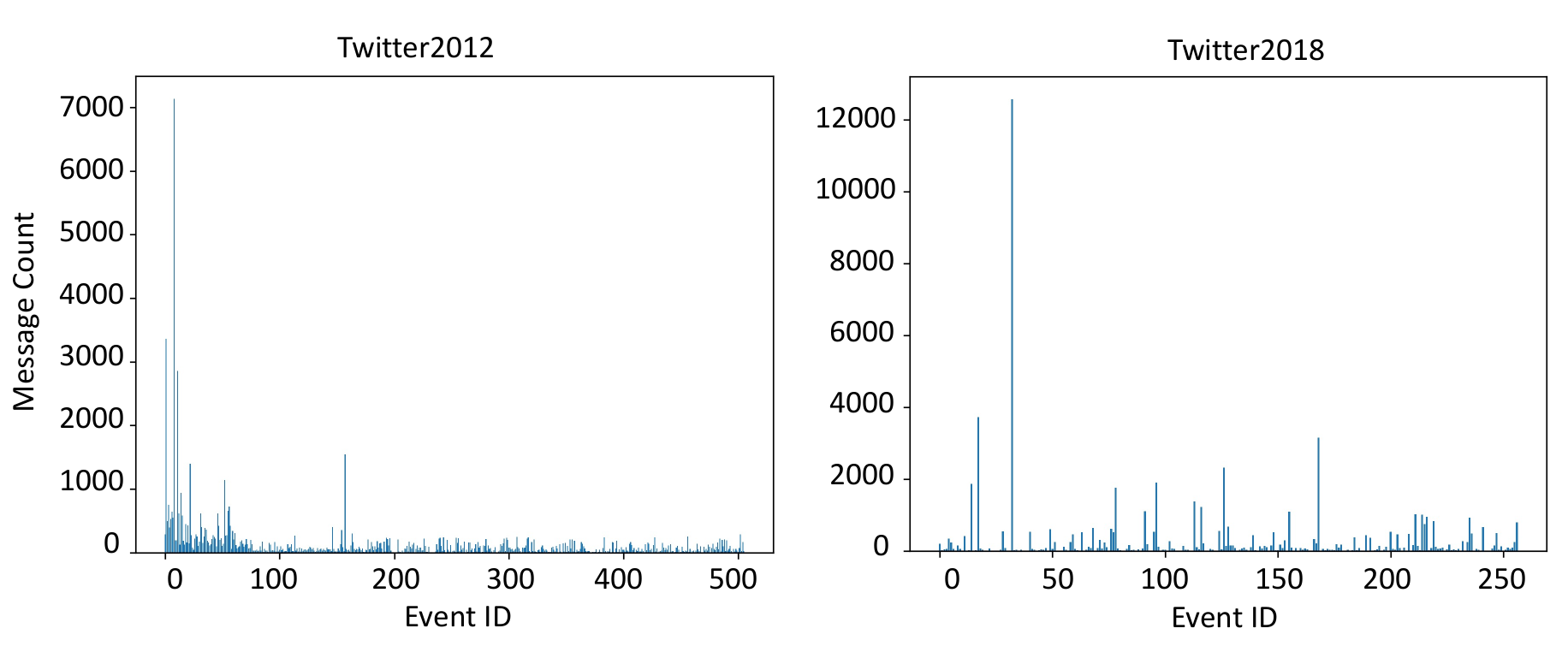}
\caption{Data imbalanced in the SED datasets. }
\vspace{-2mm}
\label{fig:data_distribution}
\end{figure}

It is worth noting that implicit augmentation methods consistently improved Macro F1 scores on the Twitter2018, increasing the baseline from 82.14\% to 86.43\%.
To understand this, we analyzed the event distribution in both the Twitter2012 and Twitter2018 datasets, as shown in Figure \ref{fig:data_distribution}. The analysis revealed that Twitter2018 has a more severe class imbalance compared to Twitter2012, with some event categories containing up to 12,000 messages. In datasets with higher class imbalance, feature space augmentation can better assist the model in learning the characteristics of rare classes, thereby improving its performance across different categories. In contrast, the more balanced Kawarith6 and Twitter2012 datasets saw less improvement in Macro F1 scores, indicating that implicit augmentation is particularly effective for underrepresented classes.
These methods directly transform structural-fused message embeddings, capturing the semantic and structural information in the data, which strengthens the model’s performance across different classes and reliefs the challenges of class imbalance.

\begin{table}[h]
\setlength\tabcolsep{3pt}
    \centering
    \caption{Different implicit augmentation methods, with all results presented as percentages.}
     \vspace{-2mm}
    \scalebox{0.62}{\begin{tabular}{@{}l|cc|cc|cc@{}}
\hline
\hline
Datasets & \multicolumn{2}{c|}{Kawarith6} & \multicolumn{2}{c|}{Twitter2012} & \multicolumn{2}{c}{Twitter2018} \\ \hline
Methods  & Micro F1      & Macro F1      & Micro F1       & Macro F1       & Micro F1       & Macro F1       \\\hline
GP   & 98.30  & 98.21 & 92.91 & 89.47 & 88.86 & 85.47 \\
PGP  & \textbf{98.41} & \textbf{98.29} & 92.85 & 89.65 & \textbf{89.61} & \textbf{86.43} \\
IDGP & 98.30  & 98.15 & 93.01 & 89.58 & 88.12 & 84.30  \\
CGP  & 98.30  & 98.22 & \textbf{93.03} & 89.53 & 88.44 & 85.02 \\
FDP  & 98.25 & 98.13 & 92.89 &\textbf{ 89.96} & 88.43 & 84.32 \\ \hline
\hline
\end{tabular}}
\label{tab:Implicit_aug}
\end{table}

\begin{table}[h]
\setlength\tabcolsep{3pt}
    \centering
    \caption{Different extraction methods, with results in percentages. KG denotes knowledge graph.}
    \vspace{-2mm}
    \scalebox{0.62}{\begin{tabular}{@{}l|cc|cc|cc@{}}
\hline
\hline
Datasets & \multicolumn{2}{c|}{Kawarith6} & \multicolumn{2}{c|}{Twitter2012} & \multicolumn{2}{c}{Twitter2018} \\ \hline
Information     & Micro F1 & Macro F1 & Micro F1 & Macro F1 & Micro F1 & Macro F1 \\ \hline
Keywords        & \textbf{97.99}    & \textbf{97.84 }   & \textbf{92.34}    & \textbf{88.65}    & 87.52    & 80.30    \\
Entities        & 97.58    & 97.33    & 92.15    & 88.21    & \textbf{88.47}    & \textbf{81.57}    \\
KG & 97.58    & 97.32    & 91.92    & 87.10    & 87.22    & 79.57    \\ \hline \hline
\end{tabular}}
\label{tab:Information_extraction}
\end{table}

\subsection{Effectiveness of Different Types of Information for Extraction and Rewriting (Q3)}

Table \ref{tab:Information_extraction} summarizes the performance of extract and rewrite strategies using keywords, entities, and knowledge graphs. While all are valuable, keywords are generally the most effective, followed by entities and knowledge graphs.
Results indicate that using keywords extracted by the LLM yields the highest Micro F1 and Macro F1 scores across the Kawarith6 and Twitter2012 datasets. This can be attributed to the LLM's ability to identify specific terms that carry significant semantic weight in SED, helping the model focus on the most informative aspects of the message.
In the Twitter2018 dataset, the entity-based approach achieved the best scores in both Micro F1 and Macro F1, emphasizing the importance of entities like names, locations, and dates in capturing key event details. This ensures the most relevant message components are preserved, which is essential for distinguishing between different events in SED tasks.

\begin{table}
\setlength\tabcolsep{3pt}
    \centering
    \caption{Different mode used in Fourier transfer, with all results presented as percentages. }
    \vspace{-2mm}
    \scalebox{0.62}{\begin{tabular}{@{}l|cc|cc|cc@{}}
\hline
\hline
Datasets & \multicolumn{2}{c|}{Kawarith6} & \multicolumn{2}{c|}{Twitter2012} & \multicolumn{2}{c}{Twitter2018} \\ \hline
Mode & Micro F1 & Macro F1 & Micro F1 & Macro F1 & Micro F1 & Macro F1 \\ \hline
High & \textbf{98.41}    & \textbf{98.29}    & \textbf{92.89}    & \textbf{89.96 }   & \textbf{88.43}    & \textbf{84.32}    \\
Low  & 95.88    & 95.41    & 92.74    & 89.39    & 87.48    & 84.49    \\
Band & 91.15    & 91.35    & 92.94    & 89.88    & 88.10    & 84.21   \\ \hline \hline
\end{tabular}}
\label{tab:Fourier_mode}
\end{table}

\begin{table}[h]
\setlength\tabcolsep{3pt}
    \centering
    \caption{Performance on different training ratio, with all results presented as percentages.}
    \vspace{-2mm}
    \scalebox{0.62}{\begin{tabular}{@{}r|ccc|ccc|c@{}}
\hline
\hline
\multicolumn{1}{l|}{}      & \multicolumn{3}{c|}{Without augmentation} & \multicolumn{3}{c|}{With augmentation} &         \\ \hline
\multicolumn{1}{l|}{Ratio} & Micro F1     & Macro F1     & Average    & Micro F1    & Macro F1    & Average   & Improve \\ \hline
10\% & 76.55 & 61.16 & 68.86 & 82.29 & 69.59 & 75.94 & 10.29 \\
20\% & 80.75 & 70.84 & 75.80 & 87.01 & 78.40 & 82.71 & 9.12  \\
30\% & 85.65 & 75.44 & 80.55 & 87.83 & 81.97 & 84.90 & 5.41  \\
40\% & 85.75 & 76.27 & 81.01 & 89.68 & 83.67 & 86.68 & 6.99  \\
50\% & 85.50 & 78.64 & 82.07 & 91.15 & 86.17 & 88.66 & 8.03  \\
60\% & 88.50 & 82.30 & 85.40 & 90.37 & 87.05 & 88.71 & 3.88  \\
70\% & 87.72 & 83.69 & 85.71 & 93.03 & 89.53 & 91.28 & 6.50  \\ \hline \hline
\end{tabular}}
\label{tab:dataset_size}
\end{table}

\begin{figure*}[t]
\centering
\includegraphics[width=1.03\textwidth]{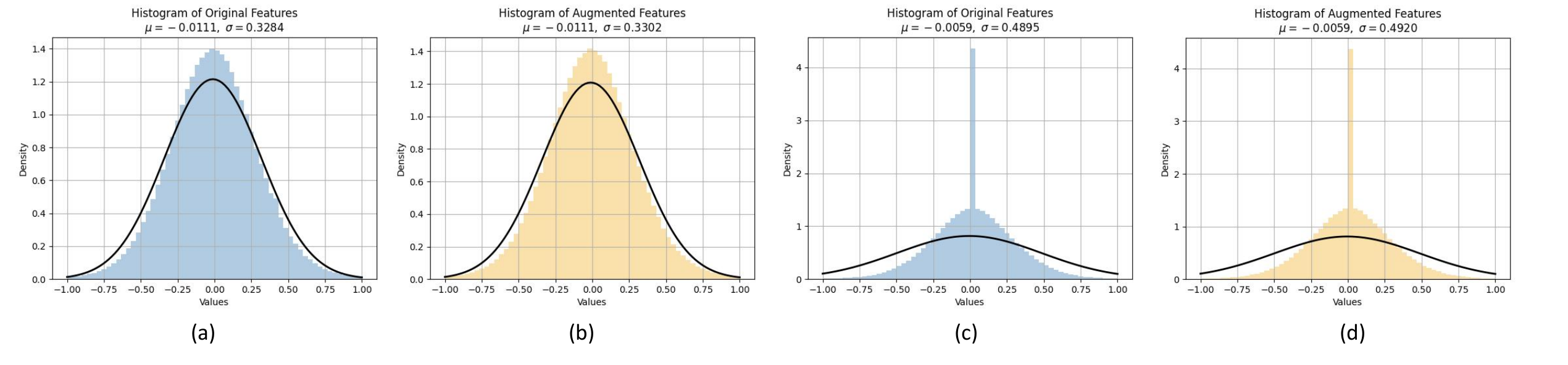}
\caption{Data histogram distribution before and after implicit augmentation (GP). Subfigure (a) and (c) are the histograms of original features before implicit augmentation; (b) and (d) are the histograms of features after augmentation.}
\label{fig:distribution-hist}
\end{figure*}

\begin{figure}[t]
\centering
\includegraphics[width=0.5\textwidth]{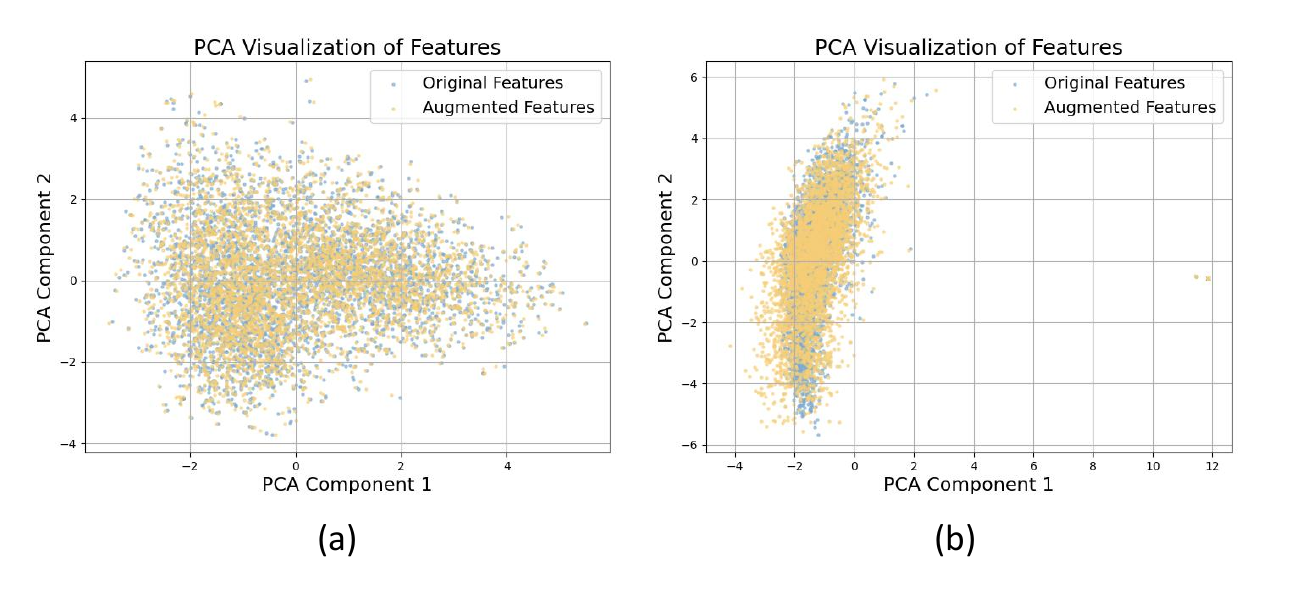}
\caption{The PCA visualization before and after implicit augmentation  (with GP noises). The blue dots in subfigures (a) and (b) represent original features; yellow dots denote augmented features.}
\label{fig:distribution-pca}
\end{figure}

\subsection{Impact of Frequency-domain Perturbations on Different Mode (Q4)}

This section investigates how different types of frequency-domain perturbations, specifically mask part of high-frequency noise, low-frequency noise, and band filtering, affect model performance across various datasets. The results, summarized in Table \ref{tab:Fourier_mode}, indicate that retain most of the high-frequency noise achieves the best results on most of cases. 
It can be attributed to its ability to retain the low-frequency components that capture essential semantic information while effectively attenuating high-frequency noise that may obscure meaningful patterns. Low-frequency components often contain critical information about the overall structure of the messages, enabling the model to maintain a robust understanding of the content. In contrast, high-frequency noise may introduce irrelevant fluctuations that hinder the model's ability to learn important features.

\subsection{Performance of SED-Aug with Limited Data (Q5)}
To evaluate the effectiveness of SED-Aug under varying data scarcity, we conducted experiments using the Twitter2012 dataset with training sets of 10\%, 20\%, ..., 70\% of the total data. We compared results with and without data augmentation, using both explicit and implicit methods. As shown in Table \ref{tab:dataset_size}, dual augmentation consistently demonstrated its utility across all data volume scenarios. Notably, our dual augmentation strategy significantly improved model performance, especially when data was limited. For example, when using only 10\% of the data, augmentation led to a performance increase of 10.29\%.
Without augmentation, performance gains diminished after 30\% of the data, stagnating at 85.40\% to 85.71\% with 60\% of the data. In contrast, dual augmentation steadily improved performance, from 88.71\% to 91.28\% with 60\% and 70\% of the data, respectively. This steady improvement can be attributed to the ability of our approach to leverage both explicit and implicit augmentation techniques, enhancing model robustness and generalization across different data settings.

\subsection{Visualizing Distribution Changes in Explicit Data Augmentation (Q6)}

We visualize the histograms of the structural fused embedding before and after implicit augmentation based on Twitter2012 dataset. In Figure \ref{fig:distribution-hist}, we find out the histogram distribution of original and augmented features keep most likely the same shapes, but slight differences, especially in variances. For subfigure (a) and (b), the mean $\mu$ of both Figures are -0.0111, but with a slight increment of variance $\sigma$ from 0.3284 to 0.3302. Similar phenomenon happens on subfigure (c) and (d). This phenomenon meets our expectations as we do not want to change the mean of the data, but just change a little bit of the features in the embedding space by adding noises (with Gaussian Perturbation noises) sampled from another Gaussian distribution to increase the sample diversity.

\needspace{1\baselineskip}
We further conduct PCA visualizations on the features before and after Gaussian Perturbation implicit augmentation (in Figure \ref{fig:distribution-pca}). Resonant with the histogram visualization, the PCA visualization also shows the general overlapping between two groups of features, but with slightly differences. This is because the implicit augmentation can increase more data diversity by adding noises in the feature space from Gaussian Perturbation.

\section{Conclusion}
In this study, we present SED-Aug, a dual data augmentation framework for SED. By combining explicit and implicit augmentation, SED-Aug enhances model performance and robustness without manual labeling.
SED-Aug significantly outperforms state-of-the-art baselines, achieving improvements of 17.67\% on the Twitter2012 dataset and 15.57\% on the Twitter2018 dataset in average F1 score. Both quantitative and qualitative experimental results show that the implicit augmentation is valuable for enhancing model robustness in imbalanced datasets and empowering data sample with more diversity, which further complements the explicit augmentation methods that are more effective in generating information-rich textual variations.

\section{Limitations}

One limitation of our work is using large language models for text augmentation lacks a clear criterion for determining the optimal amount of augmented data. While augmentation can improve model performance by enhancing data diversity, adding too much augmented data may introduce noise or redundancy. Conversely, too little augmentation may not provide sufficient variation to improve generalization. The ideal balance depends on multiple factors, including the task complexity, the quality of the original dataset, and the specific augmentation strategy used. However, there is no universally accepted guideline for how much augmentation is necessary, making it challenging to determine the best augmentation ratio.

\section*{Acknowledgement}
This work was supported by the Australian Research Council Project with No. DP230100899, Macquarie University Data Horizons Research Centre and Applied Artificial Intelligence Centre. Corresponding author: Jia Wu.

\bibliography{custom}

\appendix

\newpage
\section{Appendix}
\label{sec:appendix}

\subsection {Experimental Setting}
All datasets are divided into training, validation, and test sets, with the splits allocated in the ratio of 70\%, 10\%, and 20\%, respectively. To construct the social graph, we utilize three distinct node types: messages, users, and entities extracted from those messages. 

For generating message embeddings, we employ the BERT model, specifically the "bert-base-uncased" variant. The text node features comprise 768-dimensional embeddings derived from the textual content, which are subsequently combined with 2-dimensional embeddings that capture temporal information.
For the user node features, we integrate 768-dimensional embeddings obtained from filtered words with 2-dimensional location embeddings, ensuring a comprehensive representation of users in the graph. 

We evaluated the data quality using GPT-4o-mini, GPT-4o, and GPT-o1 on 100 examples. The results showed that GPT-4o-mini, the most cost-efficient model, produced competitively high-quality data. Consequently, we used GPT-4o-mini for all subsequent experiments, setting the maximum token limit to 1000 while keeping all other settings at their default values.
In the implicit data augmentation process for the Kawarith6 dataset, we establish a probability threshold $\alpha$ of 0.3, a standard deviation $\sigma$ of 0.01, and a clipping range $c$ of 0.005. When applying frequency-domain perturbation, the keep ratio $r$ is set to 0.98, and the noise level $\eta$ is adjusted to 0.02.
For the Twitter2012 dataset, we modify the probability threshold $\alpha$ to 0.6, the standard deviation $\sigma$ to 0.1, and the clipping range $c$ to 0.05. The frequency-domain perturbation settings for this dataset include a keep ratio $r$ of 0.95 and a noise level $\eta$ of 0.02.
Similarly, for the Twitter2018 dataset, we set the probability threshold $\alpha$ at 0.6, the standard deviation $\sigma$ at 0.1, and the clipping range $c$ at 0.0006. The frequency-domain perturbation parameters maintain a keep ratio $r$ of 0.98 and a noise level $\eta$ of 0.02.

\subsection{Datasets and Evaluation Metrics }
We conduct experiments on three datasets:  Kawarith6 \cite{alharbi2021kawarith}, Twitter2012 \cite{James2013Building}, and Twitter2018\cite{Beatrice2020a}. Kawarith6 contains 4,860 messages belonging to six unique event classes; Twitter2012 contains 68,841 messages from 503 unique event classes; Twitter2018 contains 64,516 messages from 257 unique event classes. 
For evaluation, we use Micro F1 and Macro F1. Micro F1 evaluates the overall performance in classifying instances across all classes, while Macro F1 checks each class individually before averaging these scores.
The classes refer to the predefined event categories in the dataset being used. These event classes vary depending on the specific dataset. For instance, in the Twitter2012 dataset, the classes include events such as the 2012 Nobel Prize in Literature, 2012 Presidential debates, the Bolivian radio man set on fire, etc. The exact classes depend on the annotations and categories defined within the dataset.

\subsection{Removing the explicit augmentation}

Table~\ref{tab:kawarith6_aug} presents the results of removing explicit augmentation on Kawarith6 to assess the contribution of explicit augmentation. "w/o Aug" denotes performance without any explicit augmentation. As shown, all explicit augmentation methods improve performance over the baseline, indicating their effectiveness. Among them, FDP achieves the highest Micro F1 (95.88) and Macro F1 (95.41), suggesting it is the most beneficial technique for this dataset.

\begin{table}[h]
\centering
\caption{Performance of removing the explicit augmentation methods on Kawarith6 dataset.}
\vspace{-2mm}
\scalebox{0.6}{
\begin{tabular}{l|c|c}
\hline \hline
Augmentation & Micro F1 & Macro F1 \\
\hline
GP       & 95.78 & 95.20 \\
PGP      & 95.78 & 95.39 \\
IDGP     & 95.78 & 95.36 \\
CGP      & 95.68 & 95.29 \\
FDP      & 95.88 & 95.41 \\
w/o Aug  & 95.13 & 94.58 \\
\hline \hline
\end{tabular}}
\label{tab:kawarith6_aug}
\end{table}

\begin{table}[h]\small
\setlength\tabcolsep{1pt}
\centering
\caption{Performance of different probability threshold.}
\vspace{-2mm}
\scalebox{0.85}{
\begin{tabular}{l|cc|cc|cc}
\hline \hline
 & \multicolumn{2}{c|}{Kawarith6} & \multicolumn{2}{c|}{Twitter2012} & \multicolumn{2}{c}{Twitter2018} \\
\hline
$\alpha$ & Micro F1 & Macro F1 & Micro F1 & Macro F1 & Micro F1 & Macro F1 \\
\hline
0.1 & 96.24 & 95.87 & 92.73 & 89.28 & 88.31 & 84.40 \\
0.3 & 96.35 & 95.98 & 92.97 & 89.18 & 88.78 & 85.32 \\
0.6 & 98.41 & 98.29 & 93.03 & 89.53 & 89.61 & 86.43 \\
\hline \hline
\end{tabular}}
\label{tab:threshold_comparison}
\end{table}

\subsection{Performance of different probability threshold $\alpha$ }
Table~\ref{tab:threshold_comparison} presents the model performance under different probability thresholds $\alpha$ on the Kawarith6, Twitter2012, and Twitter2018 datasets. 
In all three data sets, increasing the value of $\alpha$ leads to consistent improvements in both Micro F1 and Macro F1 scores. In the Kawarith6 dataset, when $\alpha$ is set to 0.1, the Micro F1 score is 96.24 and the Macro F1 is 95.87. 

As the threshold increases to 0.6, the scores rise substantially to 98.41 for Micro F1 and 98.29 for Macro F1, indicating a clear performance gain.
A similar pattern is observed in the Twitter2012 dataset, where increasing $\alpha$ from 0.1 to 0.6 improves Micro F1 from 92.73 to 93.03, and Macro F1 from 89.28 to 89.53. In the Twitter2018 dataset, Micro F1 increases from 88.31 at $\alpha = 0.1$ to 89.61 at $\alpha = 0.6$, while Macro F1 improves from 84.40 to 86.43.

\subsection{Evaluating LLM Performance under Zero-shot and Few-shot Settings}
We evaluated label predictions using an LLM under both zero-shot and few-shot settings on the Kawarith6 dataset. The results are as follows:

\noindent{Zero-shot: Micro F1 = 92.8\%, Macro F1 = 92.3\%};
Few-shot: Micro F1 = 98.1\%, Macro F1 = 98.0\%.

These results confirm that LLMs can achieve strong performance which are comparable to the proposed method, particularly in the few-shot setting. However, as we discussed, LLM-based approaches come with significant computational and financial costs, especially when scaling to large datasets. Our method, SED-aug, provides a more efficient alternative while maintaining better performance by including valuable graph information inside.

\begin{figure*}[t]
\centering
\includegraphics[width=0.92\textwidth]{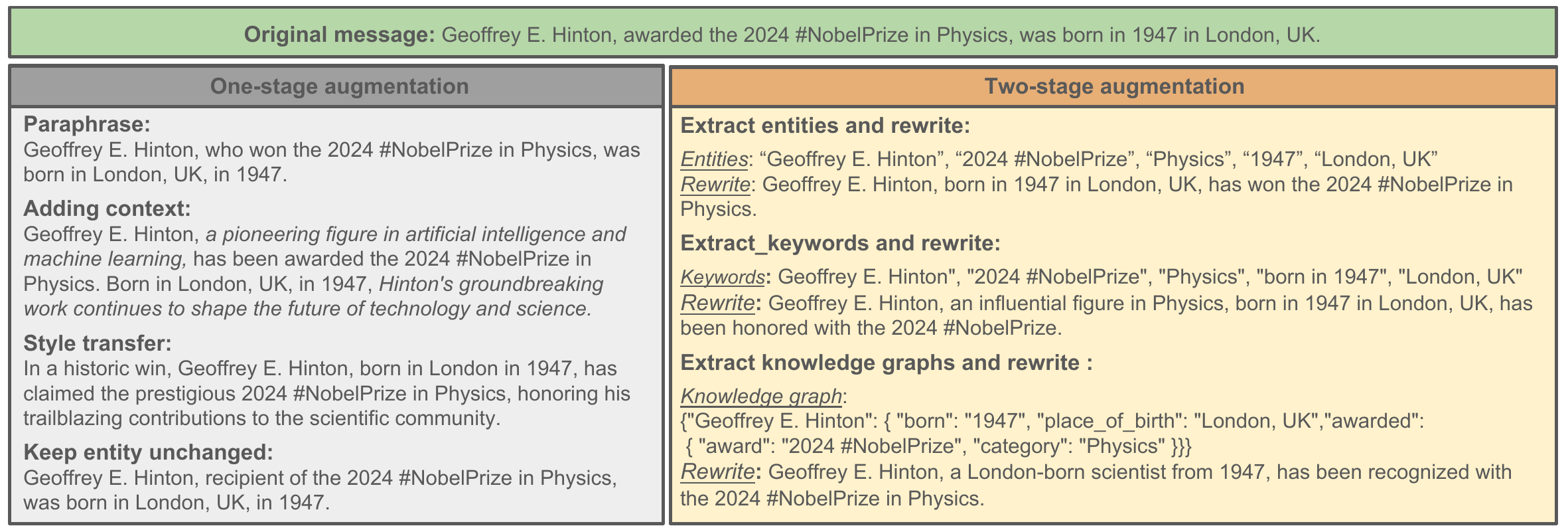}
\caption{Example of explicit augmentation. }
\vspace{-2mm}
\label{fig:example}
\end{figure*}

\begin{figure*}[h]
\centering
\includegraphics[width=0.9\textwidth]{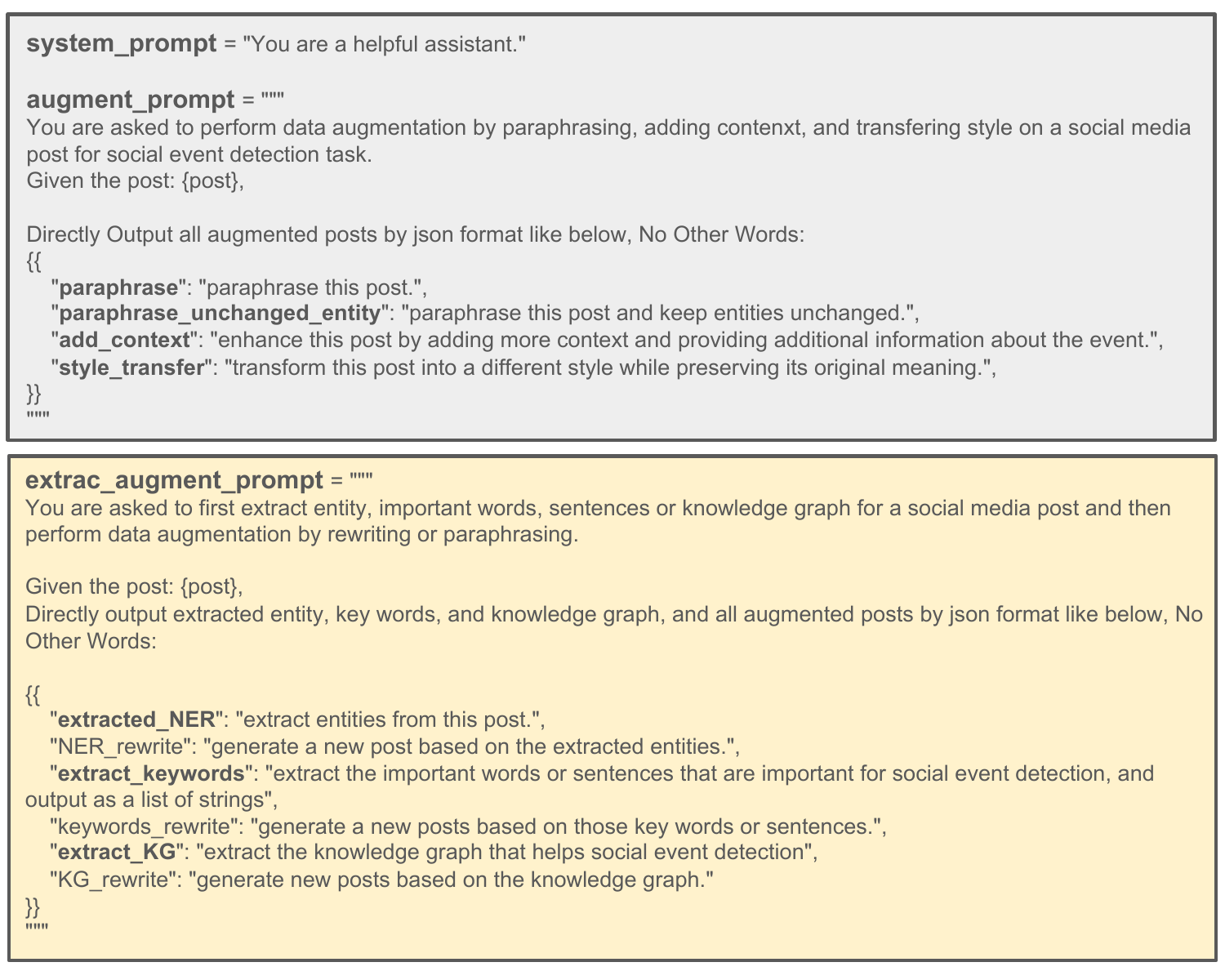}
\caption{Prompt for the explict data augmentation.}
\vspace{-2mm}
\label{fig:prompt}
\end{figure*}

\subsection{Example of explicit augmentation and prompt}

Figure \ref{fig:example} shows the example of different explicit augmentation strategies. Figure \ref{fig:prompt} is the prompt we have used for explicit augmentation.

\end{document}